\documentclass{article} 
\usepackage{iclr2027_conference,times}

\usepackage{amsmath,amsfonts,bm}

\usepackage{graphicx}
\usepackage{booktabs}
\usepackage[table]{xcolor}
\usepackage{makecell}
\usepackage{natbib} 

\usepackage{fontawesome5}
\usepackage{xcolor}

\newcommand{\modalityicon}[2]{%
  \begingroup
  \ifnum#1=1\relax
    \color{black}%
  \else
    \color{black!18}%
  \fi
  \makebox[1.15em][c]{\footnotesize #2}%
  \endgroup
}

\newcommand{\inputs}[4]{%
  \modalityicon{#1}{\textbf{\sffamily T}}%
  \modalityicon{#2}{\faImage}%
  \modalityicon{#3}{\faVideo}%
  \modalityicon{#4}{\faCube}%
}

\def\eqref#1{equation~\ref{#1}}

\def\1{\bm{1}}

\DeclareMathAlphabet{\mathsfit}{\encodingdefault}{\sfdefault}{m}{sl}
\SetMathAlphabet{\mathsfit}{bold}{\encodingdefault}{\sfdefault}{bx}{n}

\usepackage{graphicx}    
\graphicspath{{figures/}} 
\newcommand{\figplaceholder}[2][0.30\textheight]{%
  \fbox{\parbox[c][#1][c]{0.96\linewidth}{\centering\small
    \textit{Image placeholder}\\[2pt] \texttt{figures/#2}}}}
\usepackage{booktabs}    
\usepackage{amssymb}     
\usepackage{subcaption}  
\usepackage{multirow}    
\usepackage{xcolor}      
\usepackage{pgfplots}
\pgfplotsset{compat=1.18}
\usepgfplotslibrary{groupplots}
\usepackage{hyperref}
\usepackage{url}
\usepackage{tikz}
\usepackage{booktabs}
\usepackage{graphicx}
\usepackage{booktabs}
\usepackage{graphicx}
\usepackage{amssymb}
\usepackage{pifont}
\usepackage{pgfplots}
\pgfplotsset{compat=1.18}
\usepackage{xcolor}

\title{Proxy2World: Learning to Generate Worlds from Lightweight Proxies without Seeing Them}

\author{%
\hspace*{-\tabcolsep}%
\begin{tabular}[t]{@{}l@{}}
\textbf{Hongli Xu\hspace{0.8em}Weilong Yan\hspace{0.8em}Anbang Wang\hspace{0.8em}Chunyu Zou\hspace{0.8em}Siyu Hong\hspace{0.8em}Jingwei Huang}\textsuperscript{*}\\[7pt]
\normalfont Tencent \qquad \textsuperscript{*}Corresponding author.
\end{tabular}%
}

\iclrfinalcopy  

\begin{document}
\maketitle
\begin{abstract}
Lightweight scene proxies let creators control scene layout and motion
while leaving room for imagination in appearance, lighting, and visual effects. 
However, a suitable proxy is not uniquely defined,
making paired proxy--video data difficult to construct automatically at scale.
We present Proxy2World, a controllable world model that learns these complementary capabilities from ordinary posed RGBD videos, without training on authored proxy–video pairs.
The model jointly learns depth-conditioned RGB generation and
joint RGBD generation through cross-modal flow matching.
Learning both tasks enables proxy--camera hybrid denoising at inference
to follow the proxy structure while producing natural, detailed visuals.
We further introduce ProxyBench to evaluate this capability across
a diverse set of scenes, camera trajectories, and subject motions.
Experiments on ProxyBench show that Proxy2World achieves a better
balance between structural adherence and visual quality than
camera-controlled and geometry-conditioned methods, supported by
quantitative metrics, VLM assessments, human evaluations and diverse qualitative results.
\textit{Project page: }\url{https://dumdumgura.github.io/proxy2world/}

\end{abstract}

\begin{figure}[h]
  \centering
  \IfFileExists{figures/Proxy2World_Teaser_compressed.pdf}{%
    \includegraphics[width=\linewidth]{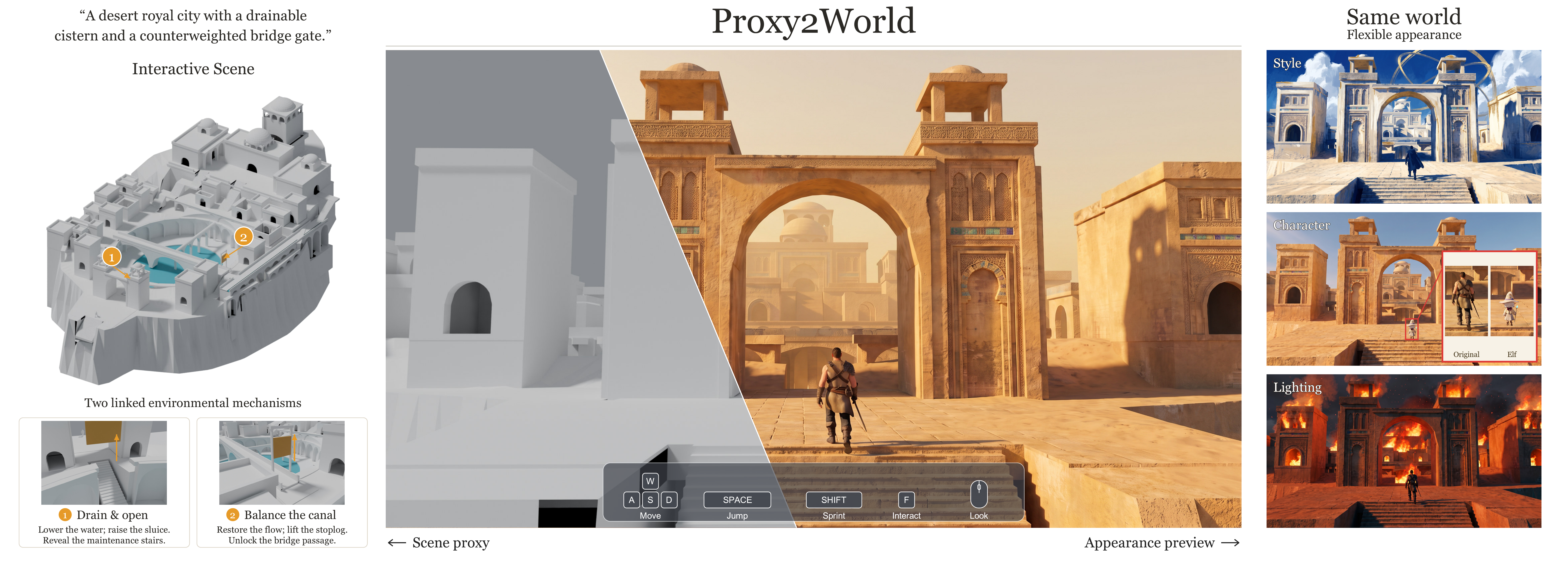}%
  }{%
    \figplaceholder{Proxy2World Teaser}%
  }
  \caption{
    \textbf{Build the structure. Generate the world.}
    Proxy2World transforms lightweight, interactive scene proxies into
    visually rich, structure-aligned worlds without paired
    proxy--video training data.
    Creators specify layouts and interactions through editable proxies
    (left), while the model generates detailed visuals (center)
    and variations in style, character, and lighting (right).
  }
  \label{fig:teaser}
\end{figure}

\section{Introduction}
\label{sec:intro}
Creating worlds that users can explore and direct is a central goal in gaming, filmmaking, and embodied AI. Recent models support action-driven interaction and camera-controlled exploration~\citep{genie,gamegenx,cameractrl2,lyra2}, while generative rendering translates explicit scene states into realistic imagery~\citep{das,diffusionrenderer,coarse2real,chen2026videomodelsnative4d}. For practical use, visual realism must be accompanied by control over scene layout, camera motion, and object movement. Yet specifying these properties should not require constructing every geometric and visual detail of the final world.

Existing approaches provide different levels of spatial control.
Camera-controlled video models, including CameraCtrl, SCoPE, and GEN3C, guide viewpoint changes through pose conditioning, ray-based representations, or reconstructed scene caches~\citep{he2024cameractrl,scope,ren2025gen3c}.
These methods enable controlled exploration, but a camera trajectory alone does not specify the desired scene layout or object motion.
Geometry-aware video models introduce additional spatial structure.
VACE and Cosmos-Transfer1 use depth or other spatial signals to guide video synthesis~\citep{vace,cosmos-transfer}, while DAR and UnividX use G-Buffer to connect editable scene states to generated observations~\citep{chen2026videomodelsnative4d,unividx}.
World-consistent Video Diffusion, FantasyWorld, and Gen3R further couple visual generation with geometry, learning to generate appearance and spatial structure together~\citep{Zhang2024WorldconsistentVD,fantasyworld,gen3r}.
Video world models draw on learned priors to synthesize rich scenes
and dynamics from sparse inputs, while generative rendering offers
direct control through explicit scene representations.
Bringing these capabilities together would allow users to prescribe scene structure and motion while leaving their visual realization to the generative model.

This raises a fundamental question: \textit{how much of a world must be specified before generation can take over?} 
We study how lightweight scene proxies can guide world generation
without specifying every detail of the final scene.
A proxy may use low-poly meshes or simple primitives and may omit parts of the scene; the goal is to follow its intended layout and motion while generating plausible geometry and appearance beyond it.

Learning to generate from such proxies, however, presents several coupled
challenges. 
First, a scene proxy reflects a designer's
abstraction of the world---which objects should be represented, how strongly they should be simplified, and which geometry can safely be omitted---so large-scale paired proxy-to-RGB data are difficult to obtain automatically.

Second, structural adherence must coexist with geometric refinement~\citep{lascomp}. A primitive specifies where an object is and how much space it occupies,
but its simplified shape should be refined rather than reproduced exactly. Recent work C2R learns control using paired synthetic coarse-to-real examples~\citep{coarse2real}; we istead aim to learn from ordinary RGBD videos and transferring that control to proxies at different levels of abstraction.
Together, these challenges raise our central question:
\textit{can proxy-based control be learned without any paired proxy-to-RGB training data and adapt to different level of abstraction?}

We answer this question with Proxy2World, a controllable world model
that learns proxy-based control from ordinary posed RGBD videos
without paired proxy--video supervision.
As illustrated in Figure~\ref{fig:teaser}, it turns editable scene proxies into
visually rich worlds while allowing variations in style, character,
and lighting.
Our key insight is to jointly learn depth-conditioned RGB generation and joint RGBD generation through cross-modal flow matching. Learning both tasks enables proxy–camera hybrid denoising at inference to follow the proxy structure while producing natural, detailed visuals.
We retain camera guidance throughout and align camera translations
with the metric scale of depth to keep the two control signals consistent.
We further introduce ProxyBench to evaluate proxy adherence,
generative refinement, and camera control across diverse scenes,
proxy abstractions, camera trajectories, and subject motions.
Experiments show that Proxy2World achieves a better balance between
structural adherence and visual quality than camera-controlled
and geometry-conditioned methods, supported by operator-based metrics,
VLM assessments, and human evaluations.

Our contributions are threefold:
\begin{enumerate}
  \item \textbf{Proxy-based world generation without paired proxy supervision.}
        We introduce Proxy2World, a controllable world model that learns structural grounding and generative completion from ordinary posed RGBD videos, without paired proxy--RGB training data.
  \item \textbf{Cross-Modal hybrid flow matching.}
        We combine proxy-grounded structure formation with camera-guided joint RGBD refinement, aligning geometry and camera motion in a shared metric scale to preserve the intended layout while refining geometry and appearance.
  \item \textbf{ProxyBench and comprehensive evaluation.}
        We introduce \textbf{ProxyBench}, an agent-driven benchmark
        spanning diverse scenes, camera trajectories, subject motions,
        and proxy abstractions.
        Evaluations using operator-based metrics, VLM assessments,
        and human preferences show that Proxy2World achieves a better
        balance between structural adherence and visual quality than
        camera-controlled and geometry-conditioned baselines.
\end{enumerate}

\section{Related Work}
\label{sec:related}
Video foundations such as Stable Video Diffusion, CogVideoX, HunyuanVideo, and Wan provide generative priors for conditional synthesis~\citep{svd,cogvideox,hunyuanvideo,wan}. We review how subsequent methods expose camera, geometry, and scene-state controls, focusing on their relationship to generation from authored proxies.

\subsection{Camera-Controlled Video Generation}
Recent video diffusion models have substantially improved explicit camera control.MotionCtrl separates camera and object motion control~\citep{wang2024motionctrl}, while CameraCtrl and VD3D inject camera representations into pretrained video models~\citep{he2024cameractrl,vd3d}. CamCo and CamI2V use epipolar constraints to structure cross-frame interactions~\citep{camco,cami2v}. CameraCtrl II extends camera-driven generation to dynamic scene exploration across wider viewpoints~\citep{cameractrl2}, and SCoPE incorporates camera sightlines into diffusion-transformer attention~\citep{scope}. These methods improve how a generator follows a requested viewpoint sequence. Geometric reprojection provides a complementary control interface. ViewCrafter uses point-based scene clues for novel-view generation~\citep{viewcrafter}, and GEN3C renders a reconstructed 3D cache along target camera trajectories~\citep{ren2025gen3c}. TrajectoryCrafter combines point-cloud renders with a source video to redirect its camera path~\citep{trajectorycrafter}. CamTrol obtains training-free camera control by using 3D layout rearrangement to guide noisy latents~\citep{camtrol}. CamCtrl3D combines pose, ray, reprojection, and 3D feature conditions~\citep{camctrl3d}, while RealCam-I2V aligns camera parameters with metric scene depth and applies scene-constrained noise shaping~\citep{realcam2v}. Proxy2World also couples geometry and camera scale, but accepts an externally authored scene proxy whose layout need not be reconstructed from the reference imagery.

\subsection{Geometry-Aware Video World Models}
Geometry can guide video synthesis as an input condition or be generated
jointly with appearance. DiffusionRenderer learns rendering through G-buffers~\citep{diffusionrenderer}, while DaS and DAR synthesize video from mesh-derived controls~\citep{das,chen2026videomodelsnative4d}. VideoComposer, Control-A-Video, VACE, and Cosmos-Transfer1 support depth or other structural conditions~\citep{videocomposer,controlavideo,vace,cosmos-transfer}, complemented by trajectory control in DragNUWA, Motion-I2V, and DragAnything~\citep{dragnuwa,motioni2v,draganything}. Geometry Forcing and GeoVideo improve geometric consistency~\citep{wu2025geometryforcing,geovideo}. WVD, Aether, Voyager, FantasyWorld, Gen3R, VideoWeave, and DualCamCtrl further couple visual and geometric generation~\citep{Zhang2024WorldconsistentVD,aether,voyager,fantasyworld,gen3r,videoweave,dualcamctrl}. We combine depth-conditioned grounding with joint RGBD refinement to follow abstract proxies while allowing their geometry to evolve.

\subsection{Proxy-to-RGB generation.}
C2R learns coarse-simulation control from paired synthetic examples~\citep{coarse2real}. Concurrent work CWM and PWM separate programmable world states from visual synthesis, training their renderers on paired proxy--video or structured-control--video data~\citep{cwm,pwm}. Marionette similarly renders predicted articulated states into pose controls for RGB generation~\citep{marionette}. Proxy2World instead learns from ordinary posed RGBD videos, without paired proxy--RGB training data. Proxy--camera hybrid denoising combines structural grounding with joint geometry--appearance refinement, enabling transfer to low-poly, primitive, and incomplete proxies.

\section{Method}
\label{sec:method}
\begin{figure}[t]
  \centering
  \IfFileExists{figures/Proxy2World_Method_v11_compact.pdf}{%
    \includegraphics[width=\linewidth]{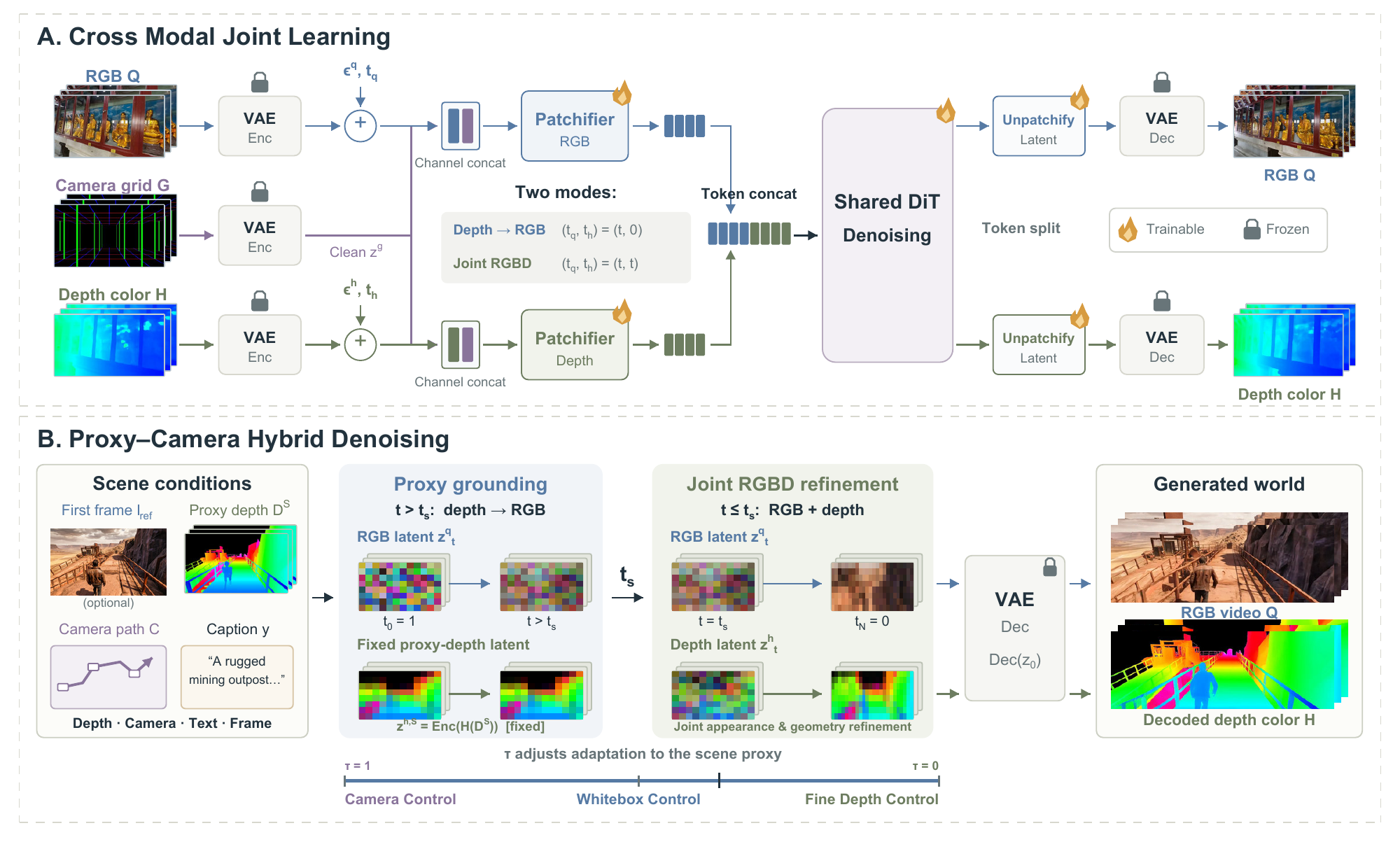}}{%
    \figplaceholder{method.png}}
  \caption{\textbf{Overview of the Method.}
        \textbf{(A) Cross Modal Joint learning.}
        We use a shared DiT to learn depth-conditioned RGB generation
        and joint RGBD generation, with camera-grid conditioning for both tasks.
        We train on ordinary posed RGBD videos without paired
        proxy--video supervision.
        \textbf{(B) Proxy--camera hybrid denoising.}
        We combine these learned capabilities during sampling:
        we first use proxy depth to establish scene structure, then
        jointly refine RGB and depth to enrich geometry and appearance.
        We retain camera-grid conditioning throughout.
        The refinement ratio $\tau$ controls the balance between
        proxy adherence and generative refinement.
        }
  \label{fig:method}
\end{figure}

\subsection{Problem Formulation}
\label{sec:problem}
Given an interactive scene proxy $S$, a control sequence $\mathcal A$,
and a camera trajectory
$\mathcal C=\{(\mathbf K_i,\mathbf R_i,\mathbf t_i)\}_{i=1}^{F}$,
we render a proxy depth sequence:
\begin{equation}
\mathbf D^{S}=\{D_i^{S}\}_{i=1}^{F}
=\mathcal R_{\mathrm{depth}}(S,\mathcal A,\mathcal C),
\end{equation}
where $F$ is the number of frames, and $\mathbf K_i$,
$\mathbf R_i$, and $\mathbf t_i$ denote camera intrinsics,
rotation, and translation, respectively.
The controls drive subject motion and scene interactions,
which are conveyed to the generative model through rendered depth.
Given a text description $y$ and an optional reference image
$\mathbf I_{\mathrm{ref}}$ spatially aligned with the initial
proxy view, our goal is to generate an RGB video:
\begin{equation}
\mathbf Q=\{\mathbf Q_i\}_{i=1}^{F}
\sim p_\theta\!\left(
\mathbf Q\mid\mathbf D^{S},\mathcal C,\mathbf I_{\mathrm{ref}},y
\right).
\end{equation}

This task requires balancing structural adherence with generative
freedom: strict conditioning can reproduce coarse proxy artifacts,
while unconstrained generation can lose the intended layout and
behavior.
We seek to preserve proxy intent while refining geometry and
appearance, learning from ordinary RGBD videos without paired
proxy-to-RGB supervision.

\subsection{Scene Proxies for Explicit World Control}
\label{sec:representation}
A scene proxy specifies control-relevant structure and behavior
without prescribing the world's final visual realization.
Its geometry may be simplified or incomplete, preserving layout,
occupancy, and subject motion while leaving visual details to
the generative prior.
We represent camera motion and scene geometry separately,
supporting camera-only generation and additional structural
control through proxy depth.

\paragraph{Camera grid.}
Following OmniDirector~\citep{omnidirector}, we represent camera
motion by rendering a fixed spatial grid $\mathcal G$ along
the prescribed trajectory:
\begin{equation}
\mathbf G
=
\mathcal R_{\mathrm{grid}}(\mathcal G,\mathcal C).
\end{equation}
The grid provides spatial references whose projected motion
expresses camera movement without specifying the target scene's
surfaces or appearance.

\paragraph{Metric depth.}
We express video depth and camera translations in consistent
metric units.
For training sequences whose depth and camera estimates share
an arbitrary scale~\citep{syn2realdepth}, we use DA3~\citep{dav3} to estimate
first-frame metric depth and align the sequence depth to this
reference.
The resulting sequence-level scale factor $s$ is applied to
both depth and camera translations:
\begin{equation}
D_i^{\mathrm{metric}}=sD_i,
\qquad
\mathbf t_i^{\mathrm{metric}}=s\mathbf t_i.
\end{equation}
This preserves their relative geometry while anchoring both
to a common physical scale.
Authored proxy depth and camera trajectories are likewise
expressed in consistent metric units.

Following Vision Banana~\citep{visionbanana}, we convert metric
depth into a three-channel false-color representation:
\begin{equation}
\mathbf H_{i}
=
\mathcal H(D_i^{\mathrm{metric}})
=
h\!\left(f(D_i^{\mathrm{metric}})\right),
\end{equation}
where $f$ nonlinearly maps metric distances to a bounded interval
and $h$ maps the result along a Hilbert-like path on the RGB cube.
The nonlinear transform allocates greater precision to nearby
geometry, while the invertible color mapping enables recovery
of metric depth.
A fixed mapping across frames and scenes preserves scale
information and provides a common representation for observed
depth and rendered proxy geometry.

\subsection{Cross-Modal Joint Learning}
\label{sec:cross_modal_fm}

We jointly train depth-to-RGB generation and joint RGBD generation
within a single flow-matching model.
Both tasks are sampled throughout training and share the same
parameters, allowing depth to serve as either a fixed geometric
condition or a generated variable.

\paragraph{Shared architecture.}
As shown in Figure~\ref{fig:method}(A), a frozen video VAE encoder
$\mathrm{Enc}$ maps RGB, metric-depth color, and camera-grid
videos to latents $\mathbf z_0^q$, $\mathbf z_0^h$, and
$\mathbf z^g$, respectively.
The subscript $0$ denotes clean data.
Each modality latent is concatenated channel-wise with the clean
camera latent and tokenized by a separate patch embedding.
RGB and depth tokens are then concatenated along the sequence
dimension and processed by a shared DiT, which predicts
modality-specific velocities:
\begin{equation}
(\mathbf v_\theta^q,\mathbf v_\theta^h)
=
\mathbf v_\theta(
\mathbf z_{t_q}^q,\mathbf z_{t_h}^h,t_q,t_h;
\mathbf z^g,\mathbf c),
\end{equation}
where $t_q$ and $t_h$ are modality-specific noise levels,
and $\mathbf c$ contains text and optional reference-frame
conditions.
After sampling, the frozen decoder $\mathrm{Dec}$ recovers
RGB and depth-color videos.
We train the patch embeddings, shared DiT, and output projections.

\paragraph{Cross-modal flow matching.}
For each modality $m\in\{q,h\}$, we construct a noisy latent
by interpolating between the clean latent $\mathbf z_0^m$
and independent standard Gaussian noise:
\begin{equation}
\mathbf z_{t_m}^m
=
(1-t_m)\mathbf z_0^m+t_m\boldsymbol\epsilon^m,
\qquad
\boldsymbol\epsilon^m\sim\mathcal N(\mathbf 0,\mathbf I),
\end{equation}
where $t_m\in[0,1]$, with $0$ denoting clean data and $1$
denoting pure noise.
The target velocity is
$\mathbf u^m=\boldsymbol\epsilon^m-\mathbf z_0^m$.
For depth-conditioned RGB generation, we set
$(t_q,t_h)=(t,0)$: depth remains clean and only RGB is supervised.
For joint RGBD generation, we set $(t_q,t_h)=(t,t)$
and supervise both modalities.
Both tasks share the same model and are optimized through:
\begin{equation}
\mathcal L
=
\mathbb E
\left[
\sum_{m\in\{q,h\}}
w_m
\left\|
\mathbf v_\theta^m
\left(
\mathbf z_{t_q}^q,\mathbf z_{t_h}^h,t_q,t_h;
\mathbf z^g,\mathbf c
\right)
-\mathbf u^m
\right\|_2^2
\right].
\end{equation}
The expectation covers task sampling, training data, noise
levels, and Gaussian noise.
We use $(w_q,w_h)=(1,0)$ for depth-conditioned RGB generation
and $(w_q,w_h)=(1,1)$ for joint RGBD generation.
This joint training allows the same model to use depth as
a fixed structural condition or generate it together with RGB,
providing the two capabilities combined during hybrid denoising.

\subsection{Proxy--Camera Hybrid Denoising}
\label{sec:hybrid_denoising}

We compose the two learned generation modes within a single
sampling trajectory.
As illustrated in Figure~\ref{fig:method}(B), early steps use
proxy depth to establish scene structure, while later steps
jointly refine RGB and depth.
Camera-grid guidance is retained throughout both stages.
Let $\tau\in[0,1]$ denote the fraction of sampling steps allocated
to joint RGBD refinement.
For a sampling schedule $1=t_0>\cdots>t_N=0$, we switch at
$k_s=\lfloor(1-\tau)N\rfloor$, with noise level $t_s=t_{k_s}$.

\paragraph{Proxy grounding.}
We encode the proxy depth-color sequence into
$\mathbf z^{h,S}=\mathrm{Enc}(\mathcal H(\mathbf D^S))$,
where $\mathcal H$ is applied frame-wise,
and initialize the RGB latent with Gaussian noise.
During the early, high-noise steps, proxy depth remains fixed
as a clean condition, and the model updates RGB using its
depth-to-RGB mode:
\begin{equation}
\frac{d\mathbf z_t^q}{dt}
=
\mathbf v_\theta^q
\left(
\mathbf z_t^q,\mathbf z^{h,S},t,0;
\mathbf z^g,\mathbf c
\right),
\qquad t>t_s.
\end{equation}
Sampling proceeds from $t=1$ toward $t=0$.
This stage grounds generation in the proxy's layout, occupied
regions, and subject motion.

\paragraph{Joint RGBD refinement.}
At $t=t_s$, we retain the current RGB latent and initialize
the depth state by adding noise to the proxy latent at the
switching noise level:
\begin{equation}
\mathbf z_{t_s}^h
=
(1-t_s)\mathbf z^{h,S}
+t_s\boldsymbol\epsilon^h,
\qquad
\boldsymbol\epsilon^h\sim\mathcal N(\mathbf 0,\mathbf I).
\end{equation}
Depth then becomes a generated variable, and both modalities
evolve under the joint RGBD mode:
\begin{equation}
\frac{d\mathbf z_t^m}{dt}
=
\mathbf v_\theta^m
\left(
\mathbf z_t^q,\mathbf z_t^h,t,t;
\mathbf z^g,\mathbf c
\right),
\qquad m\in\{q,h\},\quad t\leq t_s.
\end{equation}
The proxy is no longer imposed as a fixed depth condition,
allowing geometry and appearance to be refined together while
camera guidance continues to constrain the viewpoint sequence.

The refinement fraction $\tau$ controls the balance between
adherence and refinement.
A smaller $\tau$ retains proxy grounding for more sampling
steps, whereas a larger $\tau$ allocates more steps to joint
refinement.
At $\tau=0$, depth remains fixed throughout sampling;
at $\tau=1$, both modalities start from noise, yielding
camera-conditioned generation without proxy grounding.
Intermediate settings combine the two capabilities using
the same checkpoint, without additional proxy-specific training.

\section{Experiments}
\label{sec:experiments}

\paragraph{Implementation details.}
We initialize Proxy2World from Wan2.2-I2V-A14B~\citep{wan}
and train on 120,624 RGBD clips from DL3DV~\citep{ling2024dl3dv},
RealEstate10K~\citep{realestate10k}, and gampeplay dataset\citep{omniworld}, each with 81 frames at $480\times832$ and 16 FPS, without paired
proxy--RGB supervision.
We freeze the video VAE and text encoder and optimize patch
embeddings, DiT blocks, and output projections using AdamW
(lr $10^{-5}$, weight decay $0.01$) on 48 GPUs.
High-/low-noise experts are trained for 50,000/35,000 steps,
with equal sampling weights for depth-to-RGB and joint RGBD
tasks and uniformly sampled discrete noise levels within
each expert's shifted-flow schedule.
Depth-to-RGB supervises RGB only; joint RGBD assigns unit
loss weights to both modalities.
The encoded RGB reference and temporal mask are concatenated
to both streams, with reference dropout $0.5$.
At inference, available reference images provide
DA3-estimated metric depth for proxy alignment, with the
same scale correction applied to camera translations.
We use 40-step Euler sampling, flow shift $3.0$, and CFG $3.5$.
The joint-refinement fraction $\tau$ determines the switch
$k_s=\lfloor(1-\tau)N\rfloor$ for
$1=t_0>\cdots>t_N=0$.
We set $\tau^\star=0.925$: three geometry-conditioned steps
followed by 37 joint RGBD steps, initializing refinement
by noising proxy depth to $t_{k_s}$.

\subsection{Experimental Setup}
\label{sec:exp:setup}
\begin{figure*}[t]
    \centering
        \includegraphics[width=\linewidth]{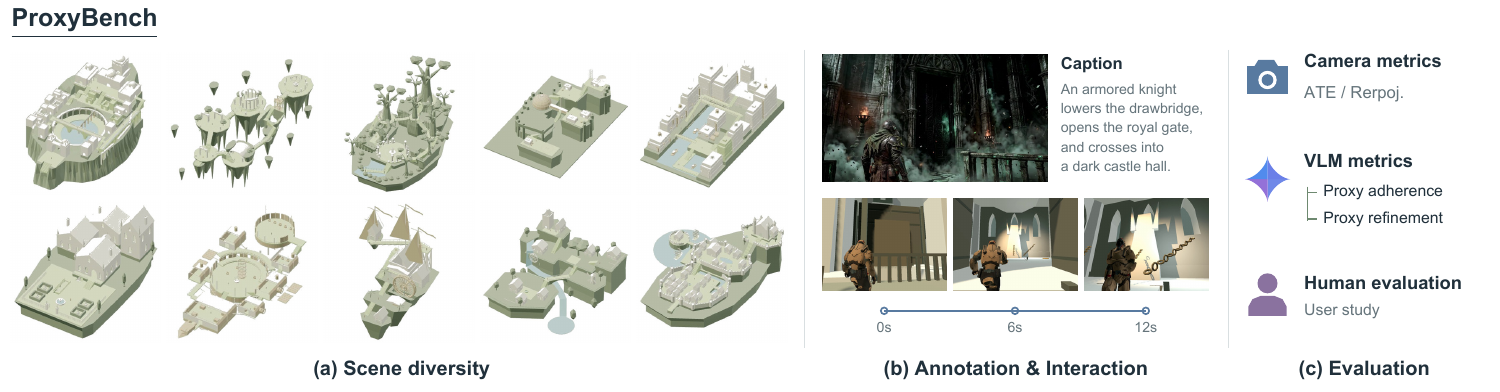}
        \caption{\textbf{}
            (a) Diverse lightweight scene proxies.
            (b) Text descriptions and scripted interactions specify the intended
            scene behavior and subject motion.
            (c) Evaluation combines camera and spatial consistency metrics,
            VLM assessments of proxy adherence and refinement, and human preferences.
        }
    \label{fig:proxybench}
\end{figure*}

\paragraph{ProxyBench.}
We introduce ProxyBench to evaluate world generation from
coarse scene proxies. It comprises 25 scenes---15 medium-scale
and 10 large-scale---with 300 five-second video sequences.
For evaluation, we select a subset of 78 sequences emphasizing
interaction and balanced coverage of motion types.
Each case provides a proxy scene, a target camera trajectory,
synchronized proxy renderings, a text prompt, and a reference
first frame. Successful generation should preserve the intended
layout and behavior while enriching coarse geometry and
appearance. Figure~\ref{fig:proxybench} provides an overview of
ProxyBench, including scene diversity, interaction annotations,
and the evaluation protocol.

\paragraph{Baselines.}
We select baselines covering geometry-conditioned rendering,
camera-controlled world generation, and coarse-to-real synthesis.
Among geometry-conditioned approaches, we focus on VACE~\citep{vace} and
Cosmos-Transfer~\citep{cosmos-transfer} as strong depth-conditioned
video generators, adapting them to proxy control using depth
rendered from our coarse scenes.
For camera-controlled generation, Lyra~2.0~\citep{lyra2}
represents projection-based approaches, but its static-scene
formulation limits dynamic interactions.
We therefore also include SCoPE/RayPE~\citep{scope} for
ray-based camera control and FantasyWorld~\citep{fantasyworld}
for joint RGBD world generation.
Coarse2Real~\citep{coarse2real} serves as the most direct
baseline, explicitly translating coarse simulations into
realistic videos.
Our camera-only ($\tau=1$), depth-only ($\tau=0$), and mixed-control
($\tau=\tau^\star$) variants share one checkpoint.
Table~\ref{tab:proxybench} specifies the inputs used by
each configuration.

\begin{table*}[t]
\centering
\caption{
Evaluation on ProxyBench.
Input icons indicate text, reference image, camera trajectory,
and proxy control; gray icons denote unused inputs.
Trajectory errors are computed from VIPE-estimated poses,
and reprojection error excludes dynamic objects.
Adherence and refinement are evaluated through Gemini pairwise
win rates (\%).
Human ranks reflect overall preference.
Bold and underline indicate the best and second-best results
within each conditioning group.
All our variants share one checkpoint trained without
paired proxy--RGB data.
}
\label{tab:proxybench}
\setlength{\tabcolsep}{3.5pt}
\renewcommand{\arraystretch}{1.12}

\resizebox{\textwidth}{!}{%
\begin{tabular}{lccccccccc}
\toprule
& 
& \multicolumn{2}{c}{Camera}
& \multicolumn{1}{c}{Spatial Cons.}
& \multicolumn{5}{c}{Proxy Evaluation} \\
\cmidrule(lr){3-4}
\cmidrule(lr){5-5}
\cmidrule(lr){6-10}

& & & &
& \multicolumn{2}{c}{Operator-based}
& \multicolumn{2}{c}{VLM-based}
& Human \\
\cmidrule(lr){6-7}
\cmidrule(lr){8-9}
\cmidrule(lr){10-10}

Method
& Input
& $\mathrm{nATE}_{t}\downarrow$
& $\mathrm{nATE}_{r}\downarrow$
& \makecell{Reproj.$\downarrow$}
& \makecell{Geo.\\Align$\uparrow$}
& \makecell{Temp.\\Align$\uparrow$}
& \makecell{Adh.\\WR$\uparrow$}
& \makecell{Ref.\\WR$\uparrow$}
& Rank$\downarrow$ \\
\midrule

\rowcolor{black!5}
\multicolumn{10}{@{}l}{%
  \small\itshape Camera-conditioned methods
} \\

Lyra 2.0~\citep{lyra2}
& \inputs{1}{1}{1}{0}
& \textbf{0.0418}
& \textbf{0.0413}
& \underline{2.171}
& \underline{0.4965}
& \textbf{0.7068}
& 32.21
& 39.48
& 10
\\

SCoPE / RAYPE~\citep{scope}
& \inputs{1}{1}{1}{0}
& 0.1066
& 0.1229
& 4.338
& 0.4508
& 0.5780
& 29.51
& 47.06
& 9
\\

FantasyWorld~\citep{fantasyworld}
& \inputs{1}{1}{1}{0}
& 0.2966
& 0.3472
& 2.449
& 0.4711
& 0.6158
& \underline{36.51}
& \underline{50.25}
& \underline{7}
\\

\rowcolor{black!3}
\textbf{Ours} (TI2V, $\tau=1$)
& \inputs{1}{1}{1}{0}
& \underline{0.0681}
& \underline{0.0524}
& \textbf{1.956}
& \textbf{0.5340}
& \underline{0.7033}
& \textbf{53.55}
& \textbf{71.44}
& \textbf{5}
\\
\midrule

\rowcolor{black!5}
\multicolumn{10}{@{}l}{%
  \small\itshape Depth-conditioned methods
} \\

VACE-Depth~\citep{vace}
& \inputs{1}{0}{0}{1}
& 0.0764
& 0.0584
& 6.264
& 0.7576
& 0.7335
& 60.14
& \textbf{45.32}
& \underline{4}
\\

Cosmos-Depth~\citep{cosmos-transfer}
& \inputs{1}{0}{0}{1}
& \underline{0.0446}
& \underline{0.0362}
& \underline{3.259}
& \underline{0.7713}
& \underline{0.7504}
& \textbf{72.26}
& \underline{40.28}
& \textbf{3}
\\

\rowcolor{black!3}
\textbf{Ours} (T2V, $\tau=0$)
& \inputs{1}{0}{0}{1}
& \textbf{0.0302}
& \textbf{0.0270}
& \textbf{2.280}
& \textbf{0.7749}
& \textbf{0.7645}
& \underline{68.79}
& 31.71
& 6
\\
\midrule

\rowcolor{black!5}
\multicolumn{10}{@{}l}{%
  \small\itshape Proxy-conditioned methods
} \\

Coarse2Real~\citep{coarse2real}
& \inputs{1}{0}{1}{1}
& 0.1637
& 0.3152
& 5.312
& 0.5520
& 0.6616
& 40.03
& 45.21
& 8
\\

\rowcolor{black!3}
\textbf{Ours} (T2V, $\tau=\tau^\star$)
& \inputs{1}{0}{1}{1}
& \underline{0.0863}
& \underline{0.0719}
& \textbf{2.067}
& \textbf{0.7084}
& \underline{0.7126}
& \underline{42.81}
& \underline{51.30}
& \underline{2}
\\

\rowcolor{black!3}
\textbf{Ours} (TI2V, $\tau=\tau^\star$)
& \inputs{1}{1}{1}{1}
& \textbf{0.0659}
& \textbf{0.0558}
& \underline{2.524}
& \underline{0.6795}
& \textbf{0.7173}
& \textbf{64.18}
& \textbf{77.96}
& \textbf{1}
\\

\bottomrule
\end{tabular}%
}
\end{table*}

\paragraph{Evaluation Protocol.}
We evaluate camera control using normalized translation
and rotation errors ($\mathrm{nATE}_{t}$ and
$\mathrm{nATE}_{r}$) computed from VIPE-estimated poses\citep{vipe}.
Spatial consistency is measured by reprojection error
after excluding dynamic objects.
Geometric and temporal alignment measure agreement with
the proxy.
We further use Gemini-v3.7-flash\citep{gemini} for pairwise evaluation of
\emph{Proxy Adherence} and \emph{Proxy Refinement}.
Adherence assesses preservation of the intended layout
and subject behavior, while refinement assesses plausible
enrichment of geometry, appearance, and motion.
We report average win rates across opponents, counting
ties as half a win, alongside overall human preference ranks. We evaluate all methods on a common set of 78 sequences using
operator-based metrics and VLM assessments, complemented by
a human preference study with 10 participants on 10 sequences.
Detailed protocols are provided in the appendix.

\begin{figure*}[t]
    \centering
        \includegraphics[width=\linewidth]{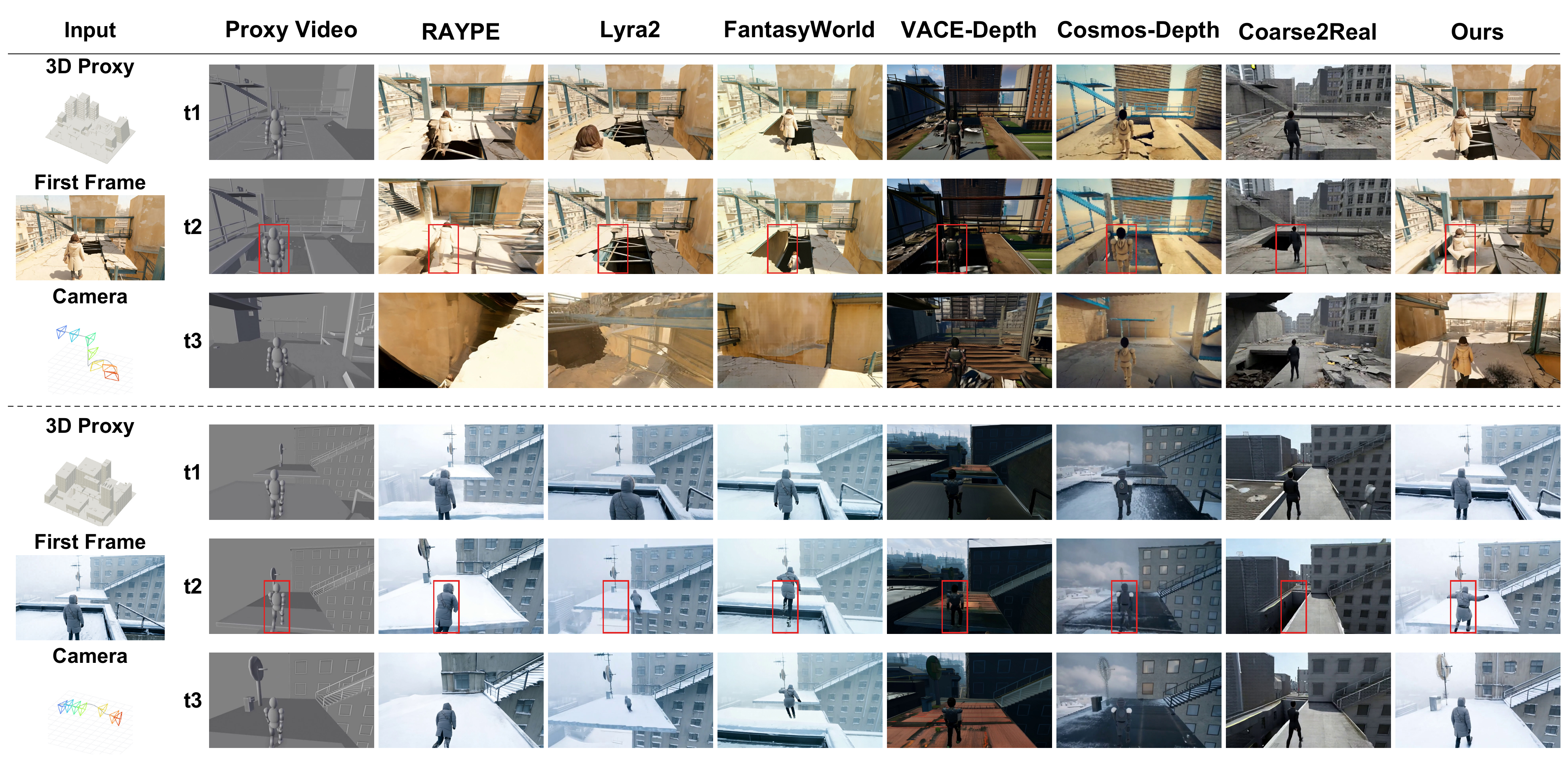}
        \caption{\textbf{Qualitative comparison on ProxyBench.}
                Red boxes highlight the intended subject locations specified by the proxy, highlighting differences in subject placement and shape. Camera-conditioned methods can deviate from the prescribed subject placement, while depth-conditioned methods can retain coarse body shapes and produced distorted characters. Ours preserves the intended placement while generating more natural character shapes and detailed scene appearance.
        }
    \label{fig:qualitative}
\end{figure*}

\subsection{Evaluation on ProxyBench}
\paragraph{Quantitative comparison.}
Table~\ref{tab:proxybench} shows that Proxy2World achieves
a strong balance between proxy adherence and generative refinement.
Compared with camera-controlled methods, our mixed TI2V model
achieves higher geometric alignment and VLM adherence while
also attaining the highest refinement win rate.
The comparison with our own camera-only TI2V variant isolates
the benefit of proxy guidance: geometric alignment improves
by 27.2\%, while adherence and refinement win rates increase
by 10.63 and 6.52 percentage points, respectively.
Thus, explicit proxy control improves structural adherence
without sacrificing the model's generative flexibility.
Compared with depth-conditioned methods, mixed generation
allows greater refinement of the supplied coarse geometry.
Cosmos-Depth achieves stronger VLM adherence
(72.26\% vs.\ 64.18\%), but our mixed TI2V model achieves
a substantially higher refinement win rate
(77.96\% vs.\ 40.28\%).
Together with its first-place ranking in human preference,
these results support the benefit of balancing structural
constraints with the freedom to refine coarse inputs.
Compared with the proxy-conditioned baseline Coarse2Real as the same proxy method,
our mixed T2V model reduces translation and rotation errors
by 47.3\% and 77.2\%, respectively, and improves geometric
alignment by 28.3\%.
These gains are obtained without authored proxy--video
training pairs, demonstrating that ordinary RGBD supervision
can support effective structural control on authored proxies.

\paragraph{Qualitative comparison.}
Figure~\ref{fig:qualitative} highlights two common failure modes.
Depth-conditioned methods follow the supplied structure but
can reproduce the coarse proxy's simplified body shapes,
resulting in distorted characters and unnatural proportions.
Camera-conditioned methods generate more natural-looking
content, but can deviate from the proxy's scene layout,
subject placement, and motion.
Proxy2World combines structural adherence with geometric
refinement: it preserves the prescribed spatial relationships
and subject motion while producing more natural character
shapes and detailed scene appearance.
The highlighted regions illustrate this difference, showing
how our method refines coarse subjects while retaining their
intended placement within the scene.

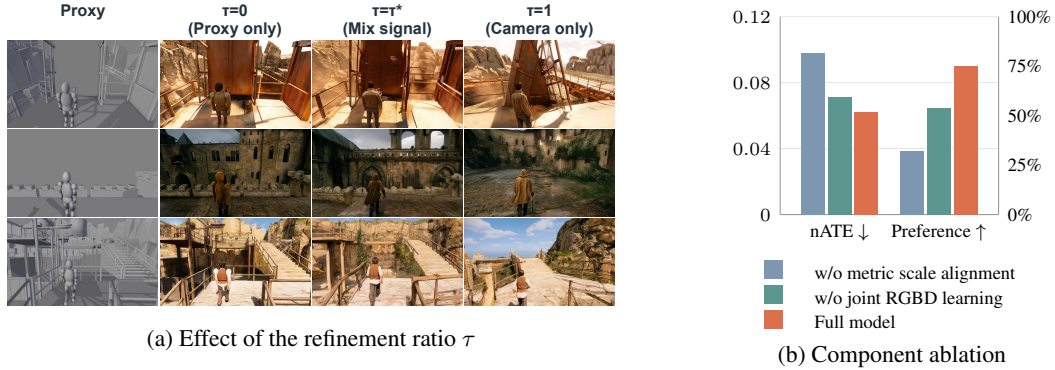
\begin{figure*}[t]
\centering

\begin{minipage}[t]{0.58\textwidth}
\vspace{0pt}
\centering

\IfFileExists{figures/ablation-selected-05-10-06-tau-compressed.pdf}
{%
    \includegraphics[width=\linewidth]{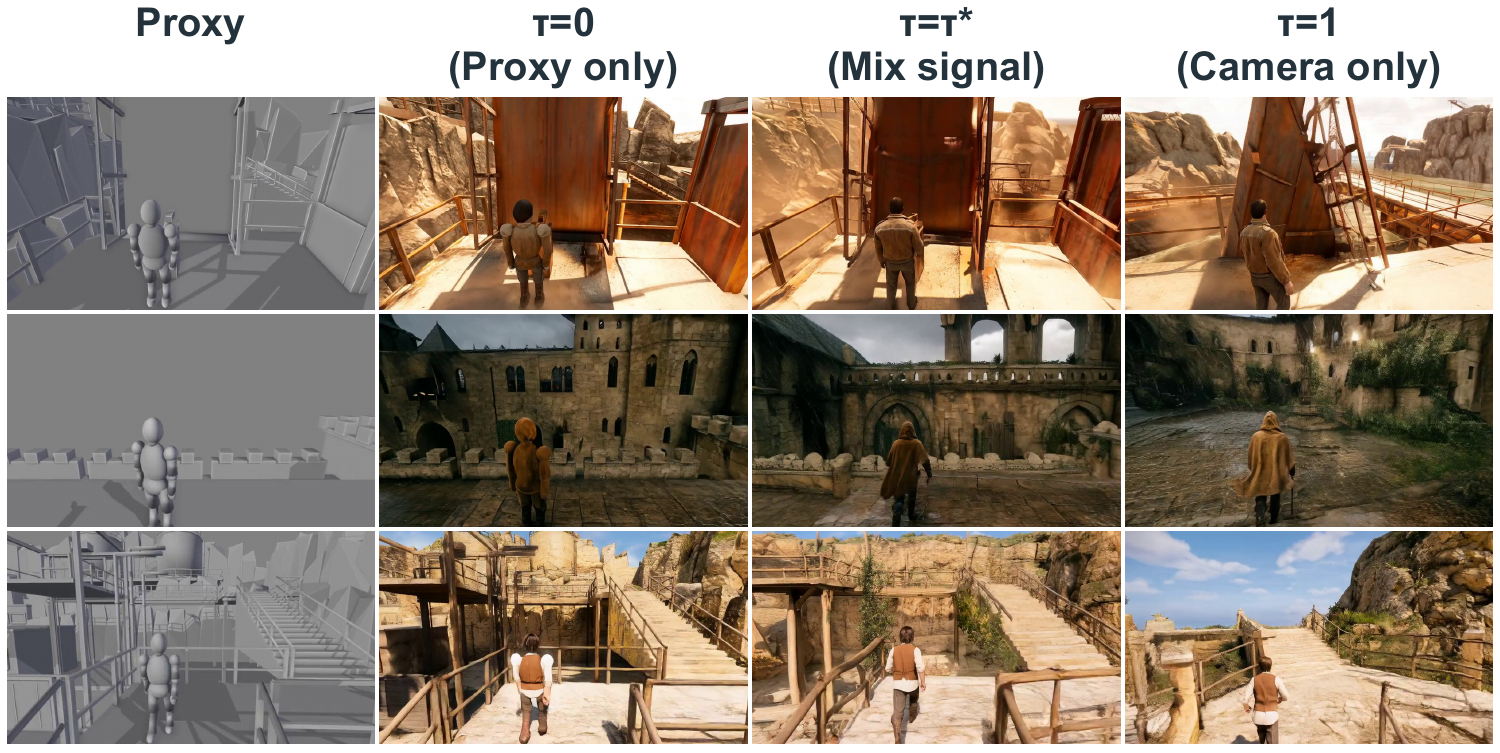}%
}{%
    \fbox{%
        \parbox[c][3.5cm][c]{0.94\linewidth}{%
            \centering
            Proxy / Reference
            \quad $\tau=0$
            \quad $\tau=\tau^\star$
            \quad $\tau=1$\\[8pt]
            Qualitative comparison
        }%
    }%
}

\smallskip
{\small (a) Effect of the refinement ratio $\tau$}
\end{minipage}
\hfill
\begin{minipage}[t]{0.32\textwidth}
\vspace{0pt}
\centering

\definecolor{ablScale}{RGB}{126,151,176}
\definecolor{ablJoint}{RGB}{92,151,143}
\definecolor{ablFull}{RGB}{220,112,76}


\begin{tikzpicture}
\pgfplotsset{
    ablpane/.style={
        width=\linewidth,
        height=4.2cm,
        xmin=0.4, xmax=2.6,
        ymin=0,
        tick label style={font=\scriptsize},
        axis line style={black!40},
        tick style={draw=none},
        scaled y ticks=false,
        bar width=9pt,
    }
}

\begin{axis}[
    ablpane,
    axis y line*=left,
    axis x line*=bottom,
    ymax=0.12,
    ytick={0,0.04,0.08,0.12},
    xtick={1,2},
    xticklabels={nATE $\downarrow$,Preference $\uparrow$},
    ymajorgrids,
    grid style={black!10},
    /pgf/number format/fixed,
    /pgf/number format/precision=2,
]
\addplot[ybar,bar shift=-10pt,fill=ablScale,draw=none]
    coordinates {(1,0.098)};
\addplot[ybar,bar shift=0pt,fill=ablJoint,draw=none]
    coordinates {(1,0.071)};
\addplot[ybar,bar shift=10pt,fill=ablFull,draw=none]
    coordinates {(1,0.062)};
\end{axis}

\begin{axis}[
    ablpane,
    axis y line*=right,
    axis x line=none,
    ymax=100,
    ytick={0,25,50,75,100},
    yticklabels={0\%,25\%,50\%,75\%,100\%},
    xtick=\empty,
]
\addplot[ybar,bar shift=-10pt,fill=ablScale,draw=none]
    coordinates {(2,32)};
\addplot[ybar,bar shift=0pt,fill=ablJoint,draw=none]
    coordinates {(2,54)};
\addplot[ybar,bar shift=10pt,fill=ablFull,draw=none]
    coordinates {(2,75)};
\end{axis}
\end{tikzpicture}

\par\smallskip
{\scriptsize
\begin{tabular}{@{}cl@{}}
\textcolor{ablScale}{\rule{7pt}{7pt}}
& w/o metric scale alignment \\
\textcolor{ablJoint}{\rule{7pt}{7pt}}
& w/o joint RGBD learning \\
\textcolor{ablFull}{\rule{7pt}{7pt}}
& Full model
\end{tabular}
}

\par\smallskip
{\small (b) Component ablation}
\end{minipage}

\caption{
Effect of the refinement ratio and model components.
(a) Varying $\tau$ with all other inputs and the random seed fixed
illustrates the trade-off between proxy adherence and refinement.
(b) Ablation of metric scale alignment and joint RGBD learning,
evaluated through camera accuracy and human preference.
}
\label{fig:ablations}
\end{figure*}

\subsection{Ablation Studies}
\label{sec:ablations}
\paragraph{Hybrid denoising.}
Figure~\ref{fig:ablations}(a) examines how the two generation
modes contribute to proxy-based control.
Proxy-only generation preserves the prescribed layout and subject
placement, but tends to carry simplified proxy shapes into the
output, particularly for characters.
Camera-only generation produces more natural shapes and richer
appearance, but can alter subject placement and scene structure.
Mixed generation preserves the major spatial relationships while
refining coarse characters and adding scene details.
These results support our use of early proxy grounding to establish
structure and subsequent joint RGBD refinement to improve its
visual realization, combining the strengths of both learned modes
without additional proxy-specific training.

\paragraph{Metric scale alignment.}
We align depth and camera translations to a common metric scale
across training scenes.
Normalizing both consistently within each scene preserves their
relative geometry, but does not establish a shared physical scale
across scenes.
Metric alignment therefore provides a consistent scale convention
for learning from depth and camera-grid conditioning.
Figure~\ref{fig:ablations}(b) shows that removing this alignment
during training yields the largest camera error among the tested
variants and substantially lowers human preference.
These results highlight the importance of a shared metric scale
across training scenes for accurate camera control and
preferred visual results.

\paragraph{Joint RGBD learning.}
To test whether refinement requires joint geometry--appearance
modeling, we replace joint RGBD generation with camera-conditioned
RGB-only generation while retaining the initial depth-conditioned
grounding stage.
This variant still releases the fixed proxy-depth constraint
during later sampling steps, but no longer generates depth
alongside RGB.
As shown in Figure~\ref{fig:ablations}(b), it yields higher camera
error and lower human preference than the full model.
The improvement therefore cannot be attributed solely to removing
the depth constraint: explicitly generating geometry together with
appearance contributes to both camera accuracy and visual quality.
This supports joint RGBD learning as a key component of our
refinement stage.

\section{Conclusion}
\label{sec:conclusion}
We presented Proxy2World, a unified RGBD world model that generates
camera-controllable videos from lightweight proxies without paired
proxy--video training data.
Proxy--camera hybrid denoising combines structural grounding with
joint RGBD refinement, balancing proxy adherence and visual quality
as demonstrated on ProxyBench.
Our approach lets creators specify coarse structure while leaving
visual details to generation.
Long-horizon interactive generation remains future work.

\bibliography{iclr2027_conference}
\bibliographystyle{iclr2027_conference}


\end{document}